\documentclass[runningheads]{llncs}

\usepackage[mobile]{eccv}

\usepackage{orcidlink}
\usepackage{float}
\usepackage{amsmath}
\usepackage{amssymb}\usepackage{amsfonts}

\usepackage[T1]{fontenc}
\usepackage{booktabs}
\usepackage[table,xcdraw]{xcolor}
\usepackage{multirow}
\usepackage[T1]{fontenc}
\usepackage{graphicx}
\usepackage{hyperref}
\usepackage{color}
\usepackage{array}
\usepackage{colortbl}   

\usepackage{amsmath, amssymb, booktabs, algorithm, algpseudocode}

\usepackage{pifont}   
\usepackage{xcolor}   

\newcommand{\cmark}{\textcolor{green}{\ding{51}}} 
\newcommand{\xmark}{\textcolor{red}{\ding{55}}}   
\usepackage{graphicx}
\usepackage{hyperref}
\usepackage{color}

\usepackage{todonotes}
\begin{document}
\title{A Lightweight Context-Driven Training-Free Network for Scene Text Segmentation and Recognition  }
\titlerunning{Context Driven STR}
%
\author{
    Ritabrata Chakraborty\inst{1,2}\orcidlink{0009-0009-3597-3703}%
    \thanks{Work done during internship at ISI Kolkata.}
    \and
    Palaiahnakote Shivakumara\inst{3}\orcidlink{0000-0001-9026-4613}
    \\
    \and
    Umapada Pal\inst{1}\orcidlink{0000-0002-5426-2618}
    \and
    Cheng-Lin Liu\inst{4}\orcidlink{0000-0002-6743-4175}
}
\institute{
    CVPR Unit, Indian Statistical Institute, Kolkata, India \\
    \email{umapada@isical.ac.in}
    \and
    Manipal University Jaipur, India \\
    \email{ritabrata.229301716@muj.manipal.edu}
    \and
    University of Salford, UK \\
    \email{s.palaiahnakote@salford.ac.uk}
    \\
    \and
    School of Artificial Intelligence, University of Chinese Academy of Sciences \\
    \email{liucl@nlpr.ia.ac.cn}
}
\authorrunning{R. Chakraborty et al.}
\maketitle              
\begin{abstract}
\vspace{-0.5cm}
Modern scene text recognition systems often depend on large end-to-end architectures that require extensive training and are prohibitively expensive for real-time scenarios. In such cases, the deployment of heavy models becomes impractical due to constraints on memory, computational resources, and latency. To address these challenges, we propose a novel, training-free plug-and-play framework that leverages the strengths of pre-trained text recognizers while minimizing redundant computations. Our approach uses context-based understanding and introduces an attention-based segmentation stage, which refines candidate text regions at the pixel level, improving downstream recognition. Instead of performing traditional text detection that follows a block-level comparison between feature map and source image and harnesses contextual information using pretrained captioners, allowing the framework to generate word predictions directly from scene context.Candidate texts are semantically and lexically evaluated to get a final score. Predictions that meet or exceed a pre-defined confidence threshold bypass the heavier process of end-to-end text STR profiling, ensuring faster inference and cutting down on unnecessary computations. Experiments on public benchmarks demonstrate that our paradigm achieves performance on par with state-of-the-art systems, yet requires substantially fewer resources.Our code can be found here: \href{https://ritabrata04.github.io/Context-driven-STR/}{https://ritabrata04.github.io/Context-driven-STR/}.
\end{abstract}
\setcounter{footnote}{0}
\section{Introduction}\label{sec:introduction}

\begin{figure}[ht]
    \centering
    \includegraphics[width=1\linewidth]{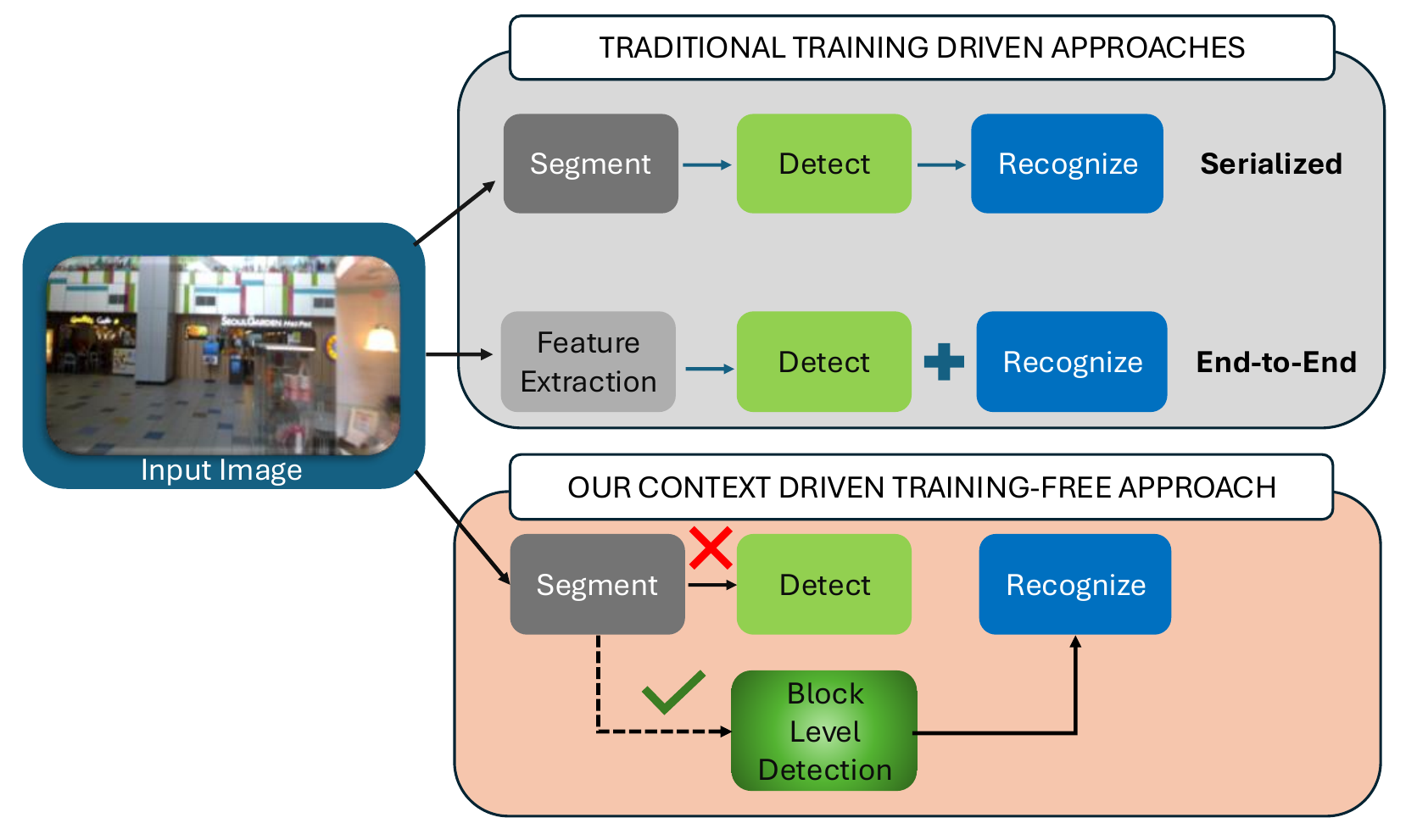}
    \caption{\textbf{A comparison of scene text recognition (STR) pipelines.} Existing works (top right) take the input image and necessarily pass it through a detection stage followed by either a sequential recognition model (Serialized), or where the recognizer is jointly trained with the detector (End-to-End). Our method (bottom right) follows the serialized steps but skips the traditional detection process, opting for a simple block-based cropping method.}
    \label{fig:pipelines}
\end{figure}

\begin{figure}[ht]
    \centering
    \includegraphics[width=1\linewidth]{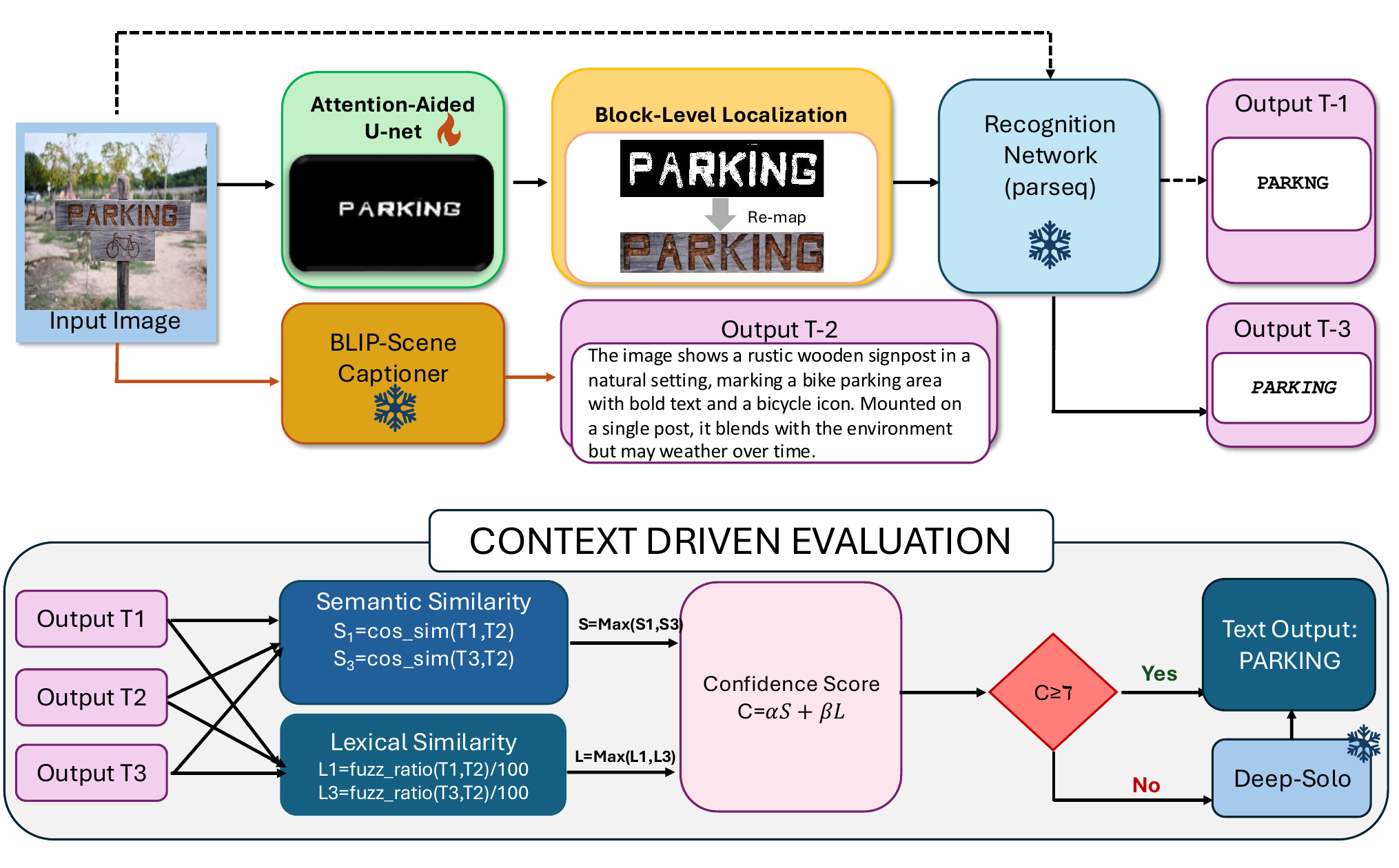}
    \caption{\textbf{Our proposed lightweight context driven network for scene text recognition.}(Top) We pass the input image to an Attention-Aided U-Net as well as a frozen image captioner. The segmentation output helps in block-level localization, which is then passed to a lightweight recognizer. This gives us two possible predictions. (Bottom) Our context driven evaluation process, with a fallback towards a heavier end-to-end recognizer (DeepSolo) uses the image caption and the two predictions to match lexical and semantic similarity to finally choose with a confidence score, which text could be the correct prediction.}
    \label{fig:architecture}
\end{figure}

The task of recognizing text in natural scene images, called Scene Text Recognition (STR) \cite{survey},  \cite{str_in_the_wild} , has garnered significant interest due to its wide range of applications in computer vision tasks \cite{str_driving}, \cite{str_robot_navigation}, \cite{str_the_deep_learning_era}. An overview of existing STR pipelines can be seen in Figure \ref{fig:pipelines}. Most state-of-the-art models \cite{deepsolo}, \cite{masktextspotterv3}, \cite{swintextspotterv2}  follow a detection-recognition paradigm, where a detection module localizes possible text instances with polygon or bounding box annotations, followed by a dedicated recognition network. Although highly effective, these models amplify the computational and memory footprint, especially in cases where heavy region-proposal based detectors are used. This leads to substantial latency and resource overhead issues, rendering their usage impractical for applications like embedded automotive modules and wearable augmented-reality devices.
\noindent To this end, we propose a lightweight context-driven training-free pipeline that aims to utilize this context in scene text recognition. Our overall pipeline, highlighted in Figure \ref{fig:architecture}, begins with an attention-based U-Net that generates rich feature maps, followed by a block-level refinement inspired by TextBlockv2 \cite{textblockv2} to get rough positional information of text. However, contrary to \cite{textblockv2}, we simply employ the binary feature maps to get pixel-based rectangular block-cuttings (cropped blocks), discussed in Section \ref{sec:methodology}. This works in conjunction with a scene captioner, which gives a textual output of image description. Our plug-and-play framework hosts a range of recognizers \cite{crnn}, \cite{parseq},  \cite{abinet}\cite{vitstr} ,which provides a prediction of the text in the localised block. Finally, we use the three text outputs : Recognizer prediction with source image \textbf{T$_{1}$}, scene description generated from captioner \textbf{T$_{2}$} and recognizer prediction with cropped instance of source image \textbf{T$_{3}$}. We employ textt comparison techniques to validate if the background cues can confidently recognize target text. If our paradigm does not yield confident prediction for a particular scenario, our pipeline defaults to a pre-trained state-of-the-art DeepSolo \cite{deepsolo}. Thus, for real-time usage where systems will have to deal with a varied range of scenes, for contextually rich cases (which we demonstrate to be the majority), our more lightweight alternative can be employed.

Overall, our contributions are threefold:
\begin{enumerate}\setlength\itemsep{0.2em}
    \item We employ off-the-shelf pre-trained models to significantly reduce both the cost and complexity arising in model training, making deployment more feasible for real-world applications.
    \item Our approach introduces an attention-gate reinforced segmentation network that pinpoints salient text regions for guiding text localization in its subsequent stages, demonstrating SOTA results across popular benchmarks.
    \item Using background cues, our proposed method can skip the heavy-weight end-to-end recognizer \cite{deepsolo} for confident predictions, reducing unnecessary computation while preserving competitive recognition performance across STR datasets.
\end{enumerate}

The remainder of this paper is organized as follows. Section \ref{sec:related works} delves into existing work and on scene text recognition frameworks, showing the need to shift from existing methods and establishing the domain gap towards our work. Section \ref{sec:methodology} provides a detailed account of our proposed framework. Experimental results, including ablation studies and efficiency analysis, are presented in Section \ref{sec: experiments}. Finally, Section \ref{sec: conclusion} concludes the paper, discussing future directions and possible improvements in our current proposal.

\section{Related Works}\label{sec:related works}

\subsection{Serialized STR frameworks}
The evolution from early scene text recognition methods to modern techniques highlights a trade-off between modular design and pattern learning quality. Two common paradigms \cite{survey} are serialized (step-by-step) \cite{serialized} and end-to-end (jointly trained) \cite{ASTER}. Classic serialized pipelines (segmentation-based recognizers) offer interpretability by explicitly locating characters and incorporating prior knowledge at multiple stages. Initial methods \cite{SDTL,RARE,ctpn} addressed issues such as irregular curved text, multiscale detection, and localization in convolutional layers. More recent frameworks like CRNN \cite{crnn} facilitate recognizing arbitrary-length text without explicit segmentation, while approaches such as \cite{kscomo} extend capabilities to multi-oriented Chinese text detection. However, serialized methods typically suffer from limited parallel processing, motivating joint supervision of detection and recognition.


\subsection{End-to-End STR Frameworks}
Early end-to-end models built upon pipeline approaches, but the advent of deep learning \cite{deep-learning} enabled jointly trained models that surpassed two-stage methods due to knowledge-sharing. Early unified models \cite{deeptextspotter,textspotting-cnn2017} employed CNN–RNN architectures for simultaneous detection and recognition. Subsequent works integrated attention mechanisms \cite{attention,Character_Region_Attention,liu2018fotsfastorientedtext}. A major breakthrough was MaskTextSpotter v1 \cite{masktextspotterv1}, inspired by Mask R-CNN \cite{he2018maskrcnn}, leveraging instance-level segmentation for curved text recognition. Later developments shifted towards one-stage, arbitrary-shape text recognition \cite{xing2019convolutionalcharacternetworks,TextDragonAE}. MaskTextSpotter v2 \cite{masktextspotterv2} used lightweight spatial attention, and MaskTextSpotter v3 \cite{masktextspotterv3} introduced an anchor-free Segmentation Proposal Network. CRAFT \cite{craft} refined detection outputs via backpropagation from recognition. Boundary-based methods \cite{boundary,textperceptron} used point-based and implicit shape supervision, respectively. Recent models prioritized towards efficiency such as \cite{pgnet,mango,pan++}.
Robust handling of curved text was enhanced by Bézier curve representations in ABCNet \cite{abcnet}, improved further in ABCNet v2 \cite{abcnetv2} with adaptive Bézier curves and positional convolutions. Transformer-based methods, such as TESTR \cite{testr}, utilized a DETR-inspired \cite{detr} dual-decoder architecture for simultaneous text detection and recognition. SwinTextSpotter \cite{swintextspotterv1,swintextspotterv2} introduced a Recognition Conversion module for feature synergy between detection and recognition tasks. Adversarial generation approaches such as TTS~\cite{banerjee2024tts} are also proving to be effective state-of-the-art ideas to explore for STR. \cite{NDorder} provided a novel way recognize text images with occluded and low-quality characters. TextTranSpotter (TTS) \cite{TTS} and SPTS \cite{SPTS} employed transformer-based direct sequence predictions.
State-of-the-art methods DeepSolo \cite{deepsolo} and DeepSolo++ \cite{deepsolo++} unified transformer decoders and explicit Bézier point queries, balancing high accuracy with model complexity which is a challenge that our work specifically addresses.

\subsection{Background Cues for STR}
\noindent A primary idea of background objects contributing towards detection of scene text is explored in \cite{Prasad_2018_ECCV} proposing a network with handcrafted modules to detect text and digits. However this does not explore the idea of using a complete scene description , limiting its scope to a limited number of object classes. Further the proposed task is restricted to detection and does not extend to scene text recognition. \cite{background_cues_in_video} explores the use of background cues in scene text videos, but does so for motional correlations between candidate text regions. Similar to \cite{Prasad_2018_ECCV}, it is restricted to text detection for multiple frames, leading to a text tracking model in video. We recognize this as a significant domain gap between existing works and our problem domain of using background context description for scene text recognition. 

\section{Methodology}\label{sec:methodology}
\subsection{Overview of Traditional STR systems} 
To understand the core components of a typical detection\=/based STR pipeline, let us first
define the problem more intuitively. Consider an input image 
\(\mathbf{I} \in \mathbb{R}^{H \times W \times 3}\), which may contain various instances 
of text amidst a potentially cluttered background. Generally, recognizing text involves two main steps: detection and recognition. Formally,
we seek a set of text regions 
\(\{\mathbf{R}_k\}_{k=1}^K\) 
such that each region \(\mathbf{R}_k\) corresponds to a bounding box (or polygon) 
enclosing text. We express the detection stage as
\(\{\mathbf{R}_k\}_{k=1}^K = \mathrm{Detect}\bigl(\mathbf{I}; \boldsymbol{\theta}_D\bigr),\)
where \(\mathrm{Detect}(\cdot; \boldsymbol{\theta}_D)\) is a function parameterized by the 
detector’s learned parameters \(\boldsymbol{\theta}_D\). This function extracts \(k\) candidate 
regions by scanning the image for salient visual cues indicative of text. Next, each region \(\mathbf{R}_k\) proceeds to the recognition step
\(\hat{\mathbf{t}}_k = \mathrm{Recognize}\bigl(\mathbf{R}_k; \boldsymbol{\theta}_R\bigr),\)
where \(\mathrm{Recognize}(\cdot; \boldsymbol{\theta}_R)\) decodes the image patch 
\(\mathbf{R}_k\) into its textual output \(\hat{\mathbf{t}}_k\), relying on parameters 
\(\boldsymbol{\theta}_R\) specific to the recognition model. While these two stages
(detection and recognition) often prove highly accurate, they come at the cost of maintaining 
and training two separate models (\(\boldsymbol{\theta}_D\) and \(\boldsymbol{\theta}_R\)).

Our work seeks to lighten the overall process described above and poses the question: \textbf{Can we reduce the computational complexity of the \(\mathrm{Detect}(\cdot; \boldsymbol{\theta}_D)\) function by "skipping" it?} To compensate for the lack of strong localization of text through bounding boxes, we seek an alternate form of improving recognition. Scene text images are innately captured with contextually relevant background objects and information. We hypothesize that \textbf{for a significant number of cases of scene text images, the context of the image can be very helpful in improving recognition accuracy. }

\subsection{Attention-Aided Segmentation}

A critical step for scene text recognition, segmentation aims to generate binary feature maps that localize features of interest at the pixel level, accurately distinguishing text (foreground) from non-text elements (background). The U\textsc{-}Net~\cite{unet} architecture is widely adopted for domain-independent segmentation tasks~\cite{text_segmentation_survey}. However, for scene text it faces two principal challenges: (i) scene images often contain diverse and cluttered backgrounds, making it difficult to separate text from non-text regions; (ii) naturally occurring text frequently appears in arbitrary orientations owing to perspective distortions, curved surfaces, or artistic placements, which can markedly degrade detection and recognition accuracy.


\begin{figure}
    \centering
    \includegraphics[width=1\linewidth]{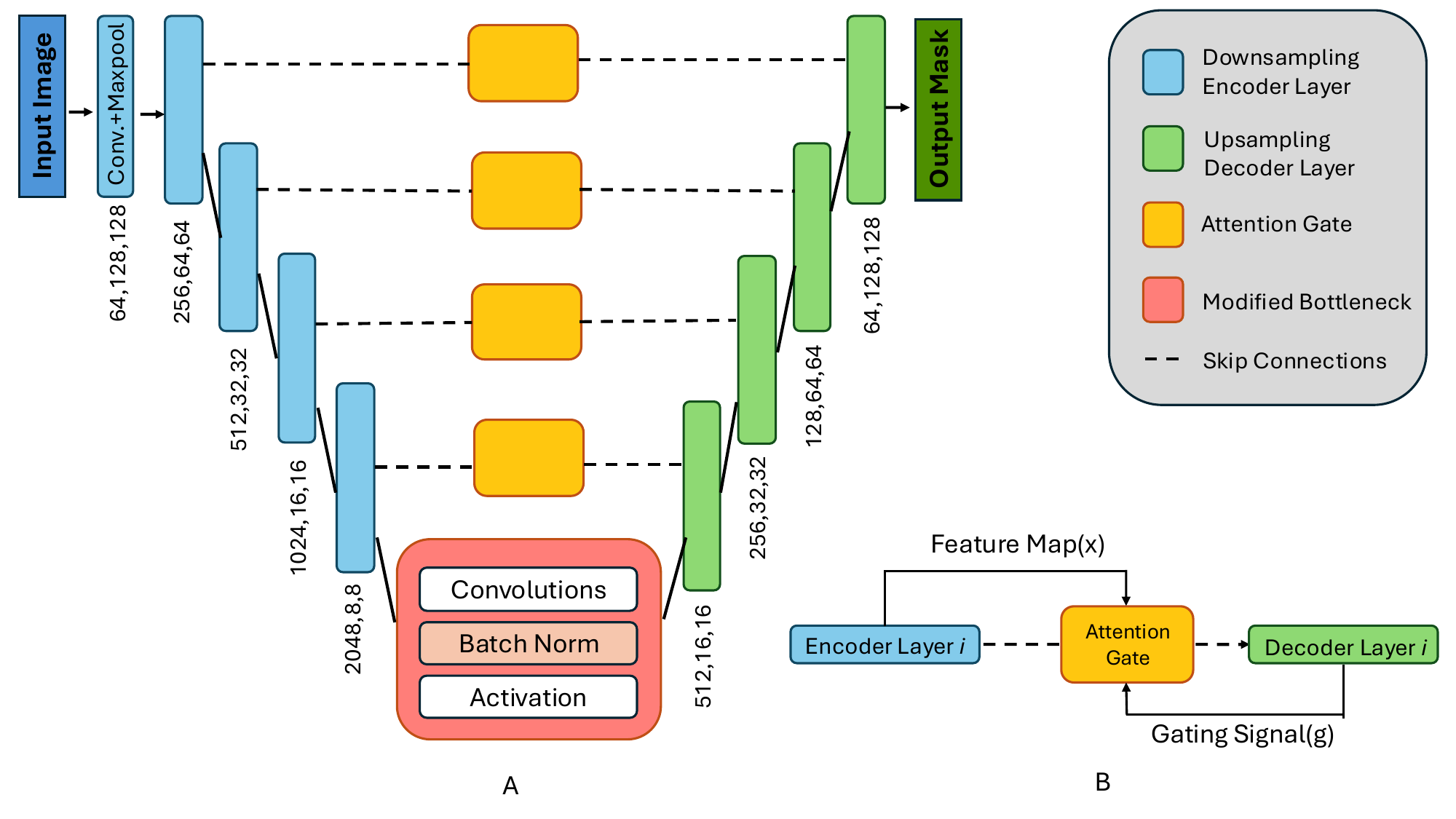}
    \caption{(Left) \textit{A:} The architecture of our modified attention-aided UNet. (Right) \textit{B:} A detailed look into the modified skip connection.}
    \label{fig:unet-mod}
\end{figure}
To overcome these issues, we design an advanced U\textsc{-}Net (Fig.~\ref{fig:unet-mod}) tailored for scene text segmentation.  
\textbf{Encoder:} We substitute the vanilla encoder with a pre\=/trained ResNet50~\cite{resnet50}, whose deep features improve robustness to complex backgrounds and diverse text patterns. 
\textbf{Skip connections:} Attention gates~\cite{attention} are inserted into each skip path, enabling the network to amplify text\=/relevant activations while suppressing background noise.  
\textbf{Bottleneck:} The deepest layer is compressed with a convolution\,+\,BatchNorm~\cite{batchnorm} block that reduces channel dimensionality, mitigates internal covariate shift, and sharpens text localization.  
Together, these modifications yield a more precise and efficient segmentation backbone, providing reliable pixel\=/level masks for downstream text localization and recognition.

\subsection{Block Level Localization}

Given a binary segmentation mask \(M \in \{0,1\}^{H \times W}\) in which text pixels are foreground, we locate each text instance by its extreme foreground coordinates, producing a bounding box \(B=(x_{\min},y_{\min},x_{\max},y_{\max})\) where, e.g., \(x_{\min}=\min\{x\mid M(x,y)=1\}\).  A padding factor \(p\) is then applied on all sides (clamped to \([0,W]\) and \([0,H]\)) before cropping the text patch \(I_{\text{crop}}=I[y_{\min}:y_{\max},x_{\min}:x_{\max}]\). When multiple text regions exist, we perform connected‐component analysis, obtaining disjoint contours \(C=\{C_i\}_{i=1}^n\) with \(C_i\cap C_j=\emptyset\).  Each contour yields a padded box \(B_i\); components with area \(A(C_i)<A_{\min}\) are discarded.  The remaining crops \(I_{\text{crop}}^{(i)}\) focus tightly on foreground text while tolerating fractional background, and serve as inputs to the recognition module. Essentially, we reduce a multiple-text instance detection requirement to single-text blocks for lightweight recognizers to work effectively. Our simple algorithm summarized in Algorithm \ref{alg:text_block_localization} thus works more efficiently eliminating the need for refined strong bounding boxes using state-of-the-art detectors such as TextFuseNet \cite{textfusenet}.  To prevent the worst-case scenario in dense text environments, we impose a manual upper bound of 10 cropped instances per image (observed in Table \ref{tab:dataset_statistics}), ensuring a controlled execution time. This makes our approach not only computationally efficient but also scalable for real-time applications, making it a superior alternative.

\begin{algorithm}[ht]
\caption{Block-Level Text Localization from Segmentation Mask}
\label{alg:text_block_localization}
\begin{algorithmic}[1]
    \Require Binary segmentation mask \( M \in \{0,1\}^{H \times W} \), original image \( I \)
    \Require Padding factor \( p \), Minimum area threshold \( A_{\min} \), Maximum text blocks \( n_{\max} = 10 \)
    \Ensure Cropped text blocks \( \{I_{\text{crop}}^{(i)}\} \) for recognition

    \State \textbf{Step 1: Extract Connected Components}
    \State \( C \gets \text{FindContours}(M) \)  \Comment{Find connected text regions in mask}
    \State \( C \gets \{C_i \mid A(C_i) \geq A_{\min} \} \)  \Comment{Filter small noisy components}
    
    \State \textbf{Step 2: Compute Bounding Boxes}
    \For{each \( C_i \) in \( C \)}
        \State Compute bounding box \( B_i = (x_{\min}^{(i)}, y_{\min}^{(i)}, x_{\max}^{(i)}, y_{\max}^{(i)}) \) 
        \State Apply padding:
        \State \( x_{\min}^{(i)} \gets \max(0, x_{\min}^{(i)} - p) \)
        \State \( y_{\min}^{(i)} \gets \max(0, y_{\min}^{(i)} - p) \)
        \State \( x_{\max}^{(i)} \gets \min(W, x_{\max}^{(i)} + p) \)
        \State \( y_{\max}^{(i)} \gets \min(H, y_{\max}^{(i)} + p) \)
    \EndFor

    \State \textbf{Step 3: Sort and Limit to Maximum Blocks}
    \State Sort \( C \) by area \( A(C_i) \) in descending order
    \State \( C \gets C[1:\min(n_{\max}, |C|)] \)  \Comment{Limit to at most \( n_{\max} = 10 \) text regions}

    \State \textbf{Step 4: Extract Text Blocks}
    \For{each \( C_i \) in \( C \)}
        \State Crop text block from original image:
        \State \( I_{\text{crop}}^{(i)} = I[y_{\min}^{(i)}: y_{\max}^{(i)}, x_{\min}^{(i)}: x_{\max}^{(i)}] \)
    \EndFor

    \State \textbf{Return} Cropped text blocks \( \{I_{\text{crop}}^{(i)}\} \)
\end{algorithmic}
\end{algorithm}


\subsection{Contextual Scene Description Generation}
\begin{figure}[ht]
    \centering
    \includegraphics[width=1\linewidth]{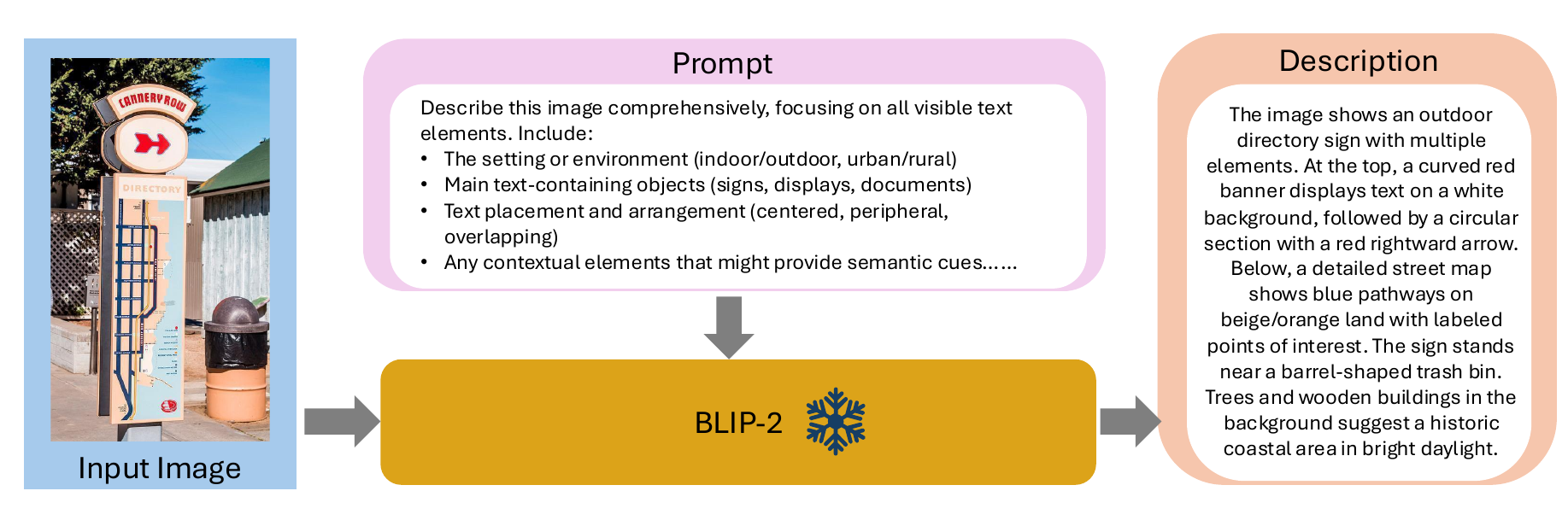}
    \caption{\textbf{Scene context generation using with image captioning.} We generate captions using BLIP2, with a detailed prompt shown above. Medium length descriptions, limited to 80 words are generated for every input image.}
    \label{fig:scene-caption}
\end{figure}
We capture background context by producing natural-language descriptions of the image with BLIP-2~\cite{BLIP2}, whose holistic outputs outperform label-only detectors such as the Google Vision API~\cite{GoogleCloudVisionAPI}.  Prompts are steered toward text-bearing surfaces, environmental cues, and possible occlusions so that the resulting descriptions directly aid recognition.To balance conciseness and detail we generate three verbosity tiers: \textit{short}, \textit{medium}, and \textit{long}, using fixed token/temperature constraints\footnote{\textit{Short}: \texttt{min\_length}=20, \texttt{max\_length}=40; \textit{medium}: 40/80; \textit{long}: 80/120}.  Empirically, the \textit{medium} tier offers the best trade-off as \textit{short} captions miss contextual cues, whereas \textit{long} ones add repetitive, text-irrelevant details. Figure \ref{fig:scene-caption} demonstrates a qualitative example of our description generation.

Through qualitative evaluation, we determine that \textit{medium-range descriptions} offer the best trade-off between completeness and relevance. These descriptions effectively capture key elements of the scene, including text-bearing surfaces (e.g., ``a storefront with a neon sign"), spatial relationships (e.g., ``a street sign mounted on a pole near a crosswalk"), and environmental cues (e.g., ``a banner hanging outside a restaurant"). Such information aids the text recognition process by contextualizing the location, likely text content, and potential occlusions, thereby reducing reliance on computationally expensive full-image processing. 

\subsection{Contextual Evaluation and Recognition Inference}
Our plug-and-play framework utilizes a diverse range of scene text recognition models, including PARSeq \cite{parseq}, CRNN \cite{crnn}, ABINet \cite{abinet}, and ViTSTR \cite{vitstr}, covering approaches from traditional recurrent architectures to transformer-based solutions. For our evaluations, we select the small variant, PARSeq-S, which offers an excellent trade-off between a lightweight design and strong recognition capabilities.

To improve framework robustness, we introduce a contextual evaluation mechanism that leverages scene descriptions. This system operates on three text representations: $T_1$, the output from the source image; $T_3$, the output from the cropped text instance; and $T_2$, a scene description generated by a pretrained BLIP2 captioning model \cite{BLIP2}. By analyzing these inputs, we reinforce accuracy by selecting the most reliable prediction.

The alignment between recognized texts ($T_1, T_3$) and the scene context ($T_2$) is assessed using two metrics, as detailed in Algorithm \ref{alg: context algorithm}. First, we measure semantic alignment using cosine similarity on sentence embeddings from the MPNet \cite{mpnet} model to derive scores $S_1$ and $S_3$. Second, we compute a lexical similarity score, $L$, using the Levenshtein fuzzy string matching ratio \cite{leventshtein_ratio}.

The final text, $T_{\text{final}}$, is initially chosen based on which candidate ($T_1$ or $T_3$) has the higher semantic similarity score with the context. We then calculate a final confidence score for this selection as a weighted combination of its semantic ($S$) and lexical ($L$) scores: $C = 0.6S + 0.4L$. If this confidence score falls below a threshold $\tau = 0.8$, the output is deemed uncertain. In these cases, the framework employs a fallback mechanism, defaulting to a pretrained DeepSolo \cite{deepsolo} recognizer to maintain high reliability. The complete procedure is formalized in Algorithm \ref{alg: context algorithm}.

\begin{algorithm}
\caption{Contextual Evaluation Algorithm}
\label{alg: context algorithm}
\begin{algorithmic}[1]
\Require Recognized texts $T_1, T_3$, Scene description $T_2$
\Ensure Final predicted text $T_{\text{final}}$

\State \textbf{Compute Similarity Scores:}
\State $S_1 \gets \text{cosine\_similarity}(\text{embedding}(T_1), \text{embedding}(T_2))$
\State $S_3 \gets \text{cosine\_similarity}(\text{embedding}(T_3), \text{embedding}(T_2))$
\State $L \gets \frac{\text{fuzz\_ratio}(T, T_2)}{100}$

\State \textbf{Compute Confidence Score:}
\State $C \gets 0.6S + 0.4L$

\State \textbf{Select Final Prediction:}
\State $T_{\text{final}} \gets T_1$ if $S_1 > S_3$ else $T_3$

\State \textbf{Handle Low Confidence:}
\If{$C < 0.8$}
    \State $T_{\text{final}} \gets \text{DeepSolo}(I)$
\EndIf

\Return $T_{\text{final}}$
\end{algorithmic}
\end{algorithm}

\section{Experiments}\label{sec: experiments}
\begin{figure}[h]
    \centering
    \includegraphics[width=1\linewidth]{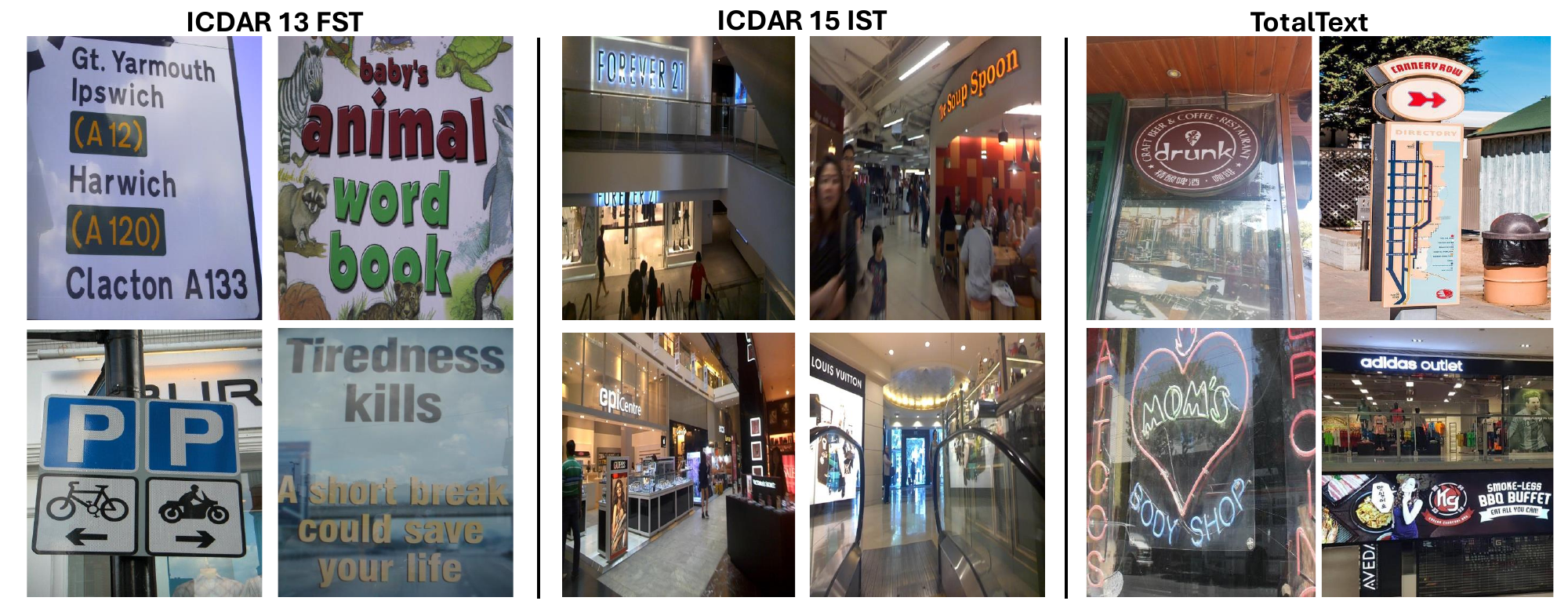}
    \caption{\textbf{Qualitative samples from datasets used for evaluation. }(Left) ICDAR13-Focused Scene Text, (Middle) ICDAR15-Incidental Scene Text, (Right) TotalText.}
    \label{fig: qual_dataset}
\end{figure}
\subsection{Datasets} In this paper, we have used 3 major scene text segmentation datasets for training and testing our attention-aided UNet: ICDAR13 FST \cite{icdar13-fst}, COCO-TS \cite{coco-ts} and Total-Text \cite{totaltext}. The datasets are statistically presented in Table \ref{tab:dataset_counts}. 
\begin{table}[h]
\centering
\caption{\textbf{Statistical details of datasets adopted for segmentation.} COCO-TS provides the highest count and diversity of text for segmentation masks.}
\label{tab:dataset_counts}
\begin{tabular}{@{}lcc@{}}
\toprule
\textbf{Dataset}       & \textbf{Training Images} & \textbf{Test Images} \\ \midrule
TotalText \cite{icdar13-fst} & 1255                    & 300                  \\
ICDAR13 FST \cite{icdar15}           & 229                     & 233                  \\
COCO-TS \cite{coco-ts}                & 43686                 & 10000                 \\
\bottomrule
\end{tabular}
\end{table}
For our overall pipeline we show recognition results on 3 standard benchmarks: ICDAR13 \cite{icdar13-fst}, ICDAR15 \cite{icdar15} and TotalText \cite{totaltext}. We show the nature of these datasets in Table \ref{tab:dataset_statistics}. A qualitative display of samples from each dataset is shown in Figure \ref{fig: qual_dataset}.
\begin{enumerate}
    \item \textbf{ICDAR13} \cite{icdar13-fst} consists of high-quality scene images (focused scene text) with horizontally aligned text. It was introduced as part of the ICDAR 2013 Robust Reading Competition and contains well-segmented word instances, making it a strong benchmark for conventional recognition models.
    \item \textbf{ICDAR15} \cite{icdar15} is a more challenging dataset featuring text captured in an incidental manner (e.g., from moving cameras). Unlike ICDAR13, this dataset contains irregularly oriented text with motion blur and occlusions, reflecting real-world recognition difficulties.
    \item \textbf{TotalText} \cite{totaltext} is designed to evaluate recognition models on arbitrarily curved text. It contains word instances with diverse orientations, making it a critical benchmark for models that need to handle non-linear text deformations effectively.
\end{enumerate}

\begin{table}[ht]
    \centering
    \caption{\textbf{Dataset statistics for overall recognition experiments.} We note the diversity of text types across the 3 datasets used in our study. H: Horizontal, MO: Multi-Oriented, C: Curved.}
    \label{tab:dataset_statistics}
    \renewcommand{\arraystretch}{1.15} 
    \begin{tabular}{lcccccc}
        \toprule
        \textbf{Datasets} & & & & \textbf{H.} & \textbf{MO} & \textbf{C} \\
        \midrule
        ICDAR-13 FST\cite{icdar13-fst} & & & & \cmark &  \xmark&  \xmark\\
        ICDAR-15 IST\cite{icdar15} & & & & \cmark & \cmark &  \xmark\\
        Total-Text\cite{totaltext} & & & & \cmark & \cmark & \cmark \\
        \bottomrule
    \end{tabular}
\end{table}

\noindent\textbf{Implementation Details.}  
Experiments are run on a single RTX\,2080 Ti (15 GB). We train a U\=/Net with Adam\cite{ADAM} (learning rate \(1\times10^{-4}\), weight decay \(0.01\)), batch size 8, and BCE loss\cite{BCELOSS}. Input images are resized to \(256\times256\); masks are down-sampled to \(128\times128\). Data augmentation uses horizontal flips and random rotations.  
At inference, batch size is set to 1 to accommodate the frozen BLIP-2 captioner\cite{BLIP2} with a ViT-B backbone\cite{vitb} (\(\sim\)120 ms/image). All other modules remain frozen, requiring only a final resize to match the recognizer’s input.

\noindent\textbf{Metrics.}  
Table~\ref{tab:segmentation_metrics} reports foreground IoU and F-score (\%). Foreground IoU is \(\text{IoU}=|P\cap G|/|P\cup G|\), where \(P\) and \(G\) are predicted and ground-truth foreground pixels. The F-score is \(F_1 = 2|P\cap G|/(|P|+|G|)\). We also measure text-recognition accuracy and FLOPs (\(10^9\)) for memory-cost analysis.
\subsection{Evaluation}
\begin{table*}[ht]
\centering
\caption{\textbf{Performance of segmentation across different datasets.} We show results for multiple baselines employing a wide variety of encoders, over 3 datasets. Our attention-guided U-Net (\colorbox{lightgray}{highlighted}) performs notably better over other models on ICDAR13 FST and COCO-TS. \textbf{Highest} and \underline{second-highest} results are marked.}
\label{tab:segmentation_metrics}

\newcommand{\thead}[1]{\begin{tabular}{@{}c@{}}#1\end{tabular}}

\small 
\setlength{\tabcolsep}{3pt} 

\begin{tabular}{@{} >{\raggedright\arraybackslash}p{3.2cm}|l|cc|cc|cc@{}}
\toprule
\textbf{Model} & \textbf{Encoder} & \multicolumn{2}{c|}{\textbf{ICDAR13 FST}} & \multicolumn{2}{c|}{\textbf{TotalText}} & \multicolumn{2}{c}{\textbf{COCO-TS}} \\ 
\cmidrule{3-8}
& & \thead{\textbf{fgIoU}\\(\%)} & \thead{\textbf{F1}\\(\%)} & \thead{\textbf{fgIoU}\\(\%)} & \thead{\textbf{F1}\\(\%)} & \thead{\textbf{fgIoU}\\(\%)} & \thead{\textbf{F1}\\(\%)} \\ 
\midrule
Vanilla U-Net (Baseline) \cite{unet} & Conv. Layers & 68.00 & 78.50 & 72.50 & 80.65 & 67.38 & 62.60 \\
DeepLabV3+ \cite{deeplabv3+} & XCeption & 69.27 & 80.20 & 74.44 & 82.42 & 72.07 & 64.10 \\ 
TextFormer \cite{textformer} & ResNet50 & 72.27 & 83.80 & \textbf{81.56} & \textbf{88.70} & \underline{73.20} & \underline{74.50}\\ 
\midrule
\rowcolor{lightgray}
\textbf{AG-UNet (Ours)} & ResNet50 & \textbf{73.14} & \textbf{85.30} & \underline{75.24} & \underline{85.44} & \textbf{75.35} & \textbf{77.50}\\ 
\bottomrule
\end{tabular}
\end{table*}

\textbf{Segmentation.} The proposed AG-UNet attains state-of-the-art fgIoU on ICDAR13 and COCO-TS and ranks second on Total-Text (behind DeepLabv3+\cite{deeplabv3+} in fgIoU and TextFormer\cite{textformer} in F1), see Table~\ref{tab:segmentation_metrics}. Relative to a vanilla UNet, it gains \(+7.71\%\) fgIoU and \(+12.6\%\) F1 on average. The larger F1 vs.\ fgIoU gap indicates recall-oriented behaviour, aligning with our block-level localisation goal of avoiding missed text pixels.

\noindent \textbf{Efficiency.} Table~\ref{tab:text_spotting_comparison} lists FLOPs for common scene-text recognizers. By avoiding multi-instance models, our context-aware pipeline attains state-of-the-art accuracy while requiring fewer operations than MaskTextSpotter\cite{masktextspotterv1}, MANGO\cite{mango}, SPTS\cite{SPTS}, TESTR\cite{testr}, and DeepSolo\cite{deepsolo}. Using DeepSolo’s ResNet backbone within our framework still cuts compute by \(\sim60\%\) relative to standalone DeepSolo.
\begin{table*}[ht]
\centering
\caption{\textbf{Efficiency comparison of STR models.} Single-instance recognizers operate on one cropped text region at a time, while end-to-end spotters detect and recognize multiple text instances jointly. Our method (\colorbox{lightgray}{highlighted}) achieves the lowest FLOPs among end-to-end spotters across both backbone configurations.}
\label{tab:text_spotting_comparison}
\renewcommand{\arraystretch}{1.0}
\setlength{\tabcolsep}{10pt}
\resizebox{\textwidth}{!}{%
\begin{tabular}{c c c r}
\toprule
\textbf{Model} & \textbf{Backbone} & \textbf{Multiple Texts?} & \textbf{FLOPs (G) $\downarrow$} \\
\midrule
\multicolumn{4}{l}{\textit{Single-instance recognizers}} \\
CRNN \cite{crnn}                         & CNN + BiLSTM            & \xmark & 2   \\
ViTSTR-S \cite{vitstr}                   & ViT                     & \xmark & 4   \\
PARSeq-S \cite{parseq}                   & Small Transformer       & \xmark & 5   \\
ABINet \cite{abinet}                     & ResNet-50 + Transformer & \xmark & 7   \\
\midrule
\multicolumn{4}{l}{\textit{End-to-end spotters}} \\
MANGO \cite{mango}                       & ResNet-50 + FPN         & \cmark & 70  \\
MaskTextSpotter \cite{masktextspotterv1} & ResNet-50 + FPN         & \cmark & 120 \\
SPTS \cite{SPTS}                         & ResNet-50 + Transformer & \cmark & 198 \\
TESTR \cite{testr}                       & ResNet-50 + Transformer & \cmark & 90  \\
DeepSolo \cite{deepsolo}                 & ResNet-50               & \cmark & 88  \\
DeepSolo \cite{deepsolo}                 & ViTAEv2-S               & \cmark & 130 \\
\midrule
\rowcolor{lightgray}\textbf{Ours} & PARSeq + ResNet-50 (DeepSolo)  & \cmark & \textbf{35} \\
\rowcolor{lightgray}\textbf{Ours} & PARSeq + ViTAEv2-S (DeepSolo) & \cmark & \textbf{50} \\
\bottomrule
\end{tabular}%
}
\end{table*}

\begin{figure}[h]
    \centering
    \includegraphics[width=\linewidth]{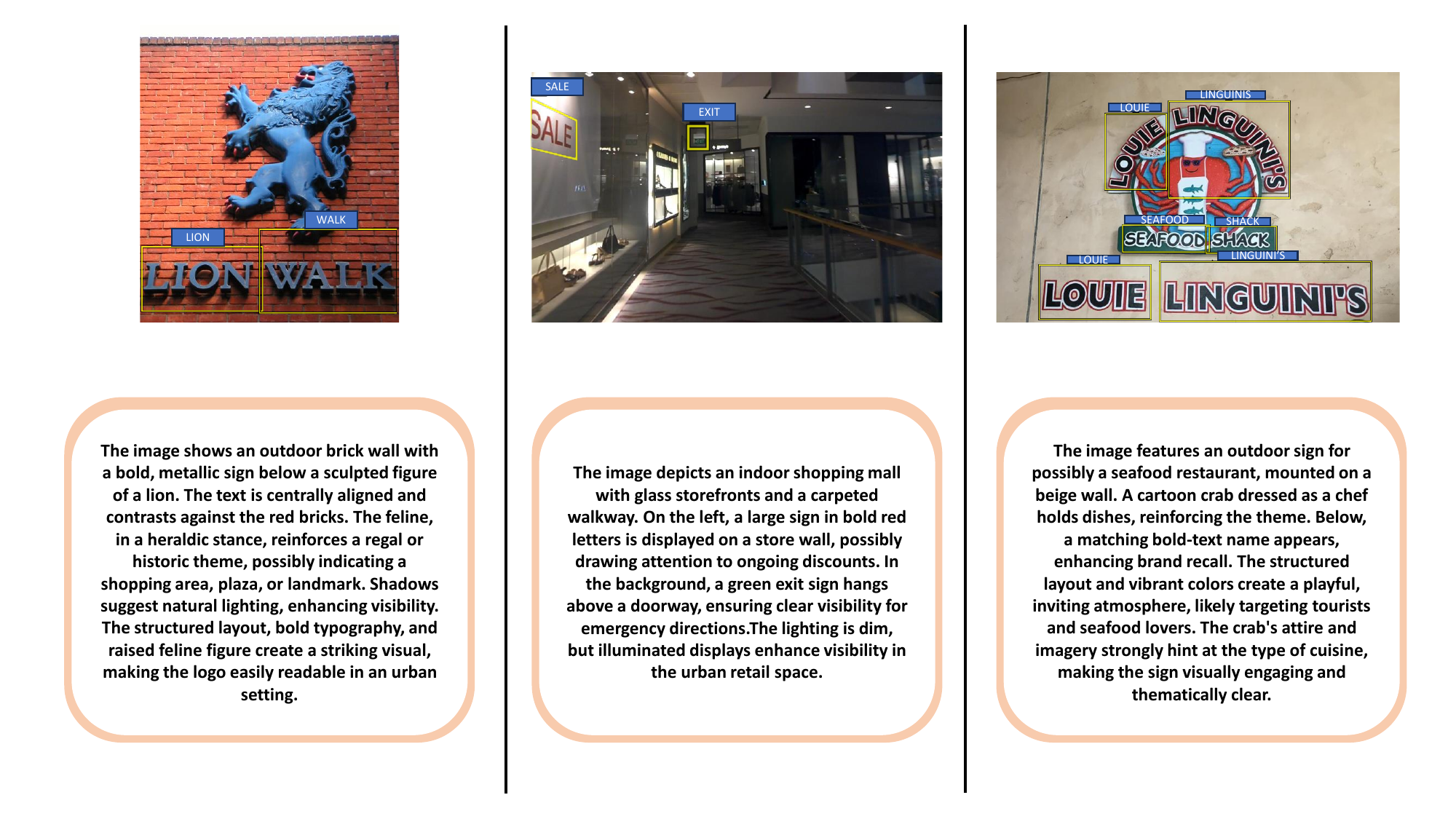}
    \caption{\textbf{Qualitative recognition results.} (Top) We show an example from each dataset(left → right): ICDAR13-FST, ICDAR15-IST, TotalText . (Bottom) We also provide the corresponding scene context caption provided by BLIP2. Best viewed in zoom.}
    \label{fig:qual_results}
\end{figure}

We evaluate recognition accuracy across the three datasets for \cite{masktextspotterv1},\cite{mango},\cite{SPTS},\cite{testr} and two versions of our proposed pipeline: PARSeq with DeepSolo having ResNet-50 \cite{resnet50} and ViTAEv2-S \cite{vitaev2} backbone. The models used for comparison are pretrained on a significant amount of  synthetic data such as \cite{synth800k} as well as real data such as \cite{scut-ctw},\cite{MLT-19},\cite{MLT-17} and \cite{textocr}. Our objective is to achieve the best or comparable to state-of-the-art accuracy while alleviating the need for excessive memory need and high FLOPs in processing recognition output. Some qualitative results are shown in Figure \ref{fig:qual_results}.
\begin{table}[t]
\centering
\caption{\textbf{Recognition accuracy (\%) across three benchmarks.} Our method (\colorbox{lightgray}{highlighted}) achieves superior performance to all baselines across all 3 datasets.}
\label{tab:merged-recog-results}
\begin{tabular}{lccc}
\toprule
\textbf{Method} & \textbf{ICDAR13-FST} & \textbf{ICDAR15-IST} & \textbf{TotalText} \\
\midrule
MaskTextSpotter\,\cite{masktextspotterv1} & 84.1 & 70.3 & 71.2 \\
MANGO\,\cite{mango}                       & 88.7 & 72.9 & 72.9 \\
SPTS\,\cite{SPTS}                         & 88.5 & 73.5 & 74.2 \\
TESTR\,\cite{testr}                       & 86.4 & 73.3 & 73.3 \\
\rowcolor{lightgray}\textbf{Ours (DeepSolo ResNet-50)} & \textbf{88.8} & \textbf{77.3} & \textbf{78.4} \\
\rowcolor{lightgray}\textbf{Ours (DeepSolo ViTAEv2-S)} & \textbf{89.7} & \textbf{78.2} & \textbf{79.8} \\
\bottomrule
\end{tabular}
\end{table}

\noindent\textbf{ICDAR13-FST.} Our pipeline surpasses serialised and end-to-end baselines, improving on MaskTextSpotter by \(+4.7\%\) (ResNet-50) and \(+5.6\%\) (ViTAEv2-S). While standalone DeepSolo peaks at 90 \%, our ViTAEv2-S variant is within 0.5 \% and the ResNet-50 variant within 1 \%. Thanks to context routing, 73 \% of images (168/229) avoid invoking DeepSolo without harming accuracy.

\noindent\textbf{ICDAR15-IST.} We obtain the best results overall, beating MaskTextSpotter by \(+7.0\%\) and \(+7.9\%\) for the two variants. Compared with DeepSolo (82 \%), the accuracy drop is below 4 \%. Although incidental-capture artefacts trigger more fallbacks, context recognition still handles 54 \% of the 500 images (272 cases).

\noindent\textbf{TotalText.} Accuracy improves by \(+7.2\%\) (ResNet-50) and \(+8.6\%\) (ViTAEv2-S) over MaskTextSpotter. Mixed-script scenes occasionally mislead the context branch, but its results trail DeepSolo by only 4.1 \% and 3.9 \%, respectively, confirming robust multilingual handling.

\begin{table}[ht]
    \centering
    \caption{Ablation Study on context scoring parameters executed on 500 test images of the ICDAR15 dataset.}
    \label{tab:ablation_study}
    \begin{tabular}{lccc}
        \toprule
        \textbf{Setting (\(\alpha, \beta, \tau\))} & \textbf{Accuracy (\%)} & \textbf{False Positives} & \textbf{Fallback to DeepSolo} \\
        \midrule
        (0.5, 0.5, 0.8) & 77.1 & 19 & 237 \\
        \rowcolor{lightgray} \textbf{(0.6, 0.4, 0.8)} & \textbf{78.2} & \textbf{13} & \textbf{228} \\
        (0.7, 0.3, 0.8) & 76.4 & 24 & 248 \\
        (0.6, 0.4, 0.75) & 78.0 & 20& 203 \\
        (0.6, 0.4, 0.85) & 77.8 & 11 & 301 \\
        \bottomrule
    \end{tabular}
    \vspace{-3pt}
\end{table}
\vspace{-6pt}
\subsection{Ablation Study}
\textbf{Context Scoring Parameter Tuning.}
An ablation over \(\alpha\!+\!\beta=1\) and decision threshold \(\tau\) (Table~\ref{tab:ablation_study}) shows that \((\alpha{=}0.6,\beta{=}0.4,\tau{=}0.8)\) offers the best trade-off: 78.2 \% accuracy, 13 false positives, and 228 DeepSolo fallbacks. Shifting to \(\alpha{=}0.7\) over-trusts context, increasing errors; \(\alpha{=}0.5\) or \(\tau{=}0.75\) lowers fallbacks but inflates false positives, while \(\tau{=}0.85\) excessively rejects predictions (301 fallbacks). These results validate that our chosen parameters provide the best balance.
\begin{table}[h]
    \centering
    \caption{\textbf{Impact of context description on recognition output.} "CBR" refers to context based recognitions, "FPR" refers to false positive recognitions.}
    \label{tab:context_impact}
    \begin{tabular}{lccc}
        \toprule
        \textbf{Scenario} & \textbf{CBR} & \textbf{Fallbacks} & \textbf{FPR} \\
        \midrule
        No Description & 163 & 337 & 23 \\
        Wrong Description & 81 & 419 & 44 \\
        \rowcolor{lightgray} \textbf{Correct Description} & \textbf{272} & \textbf{228} & \textbf{13} \\
        \bottomrule
    \end{tabular}
\end{table}
\begin{figure}[h]
    \centering
    \includegraphics[width=1.0\linewidth]{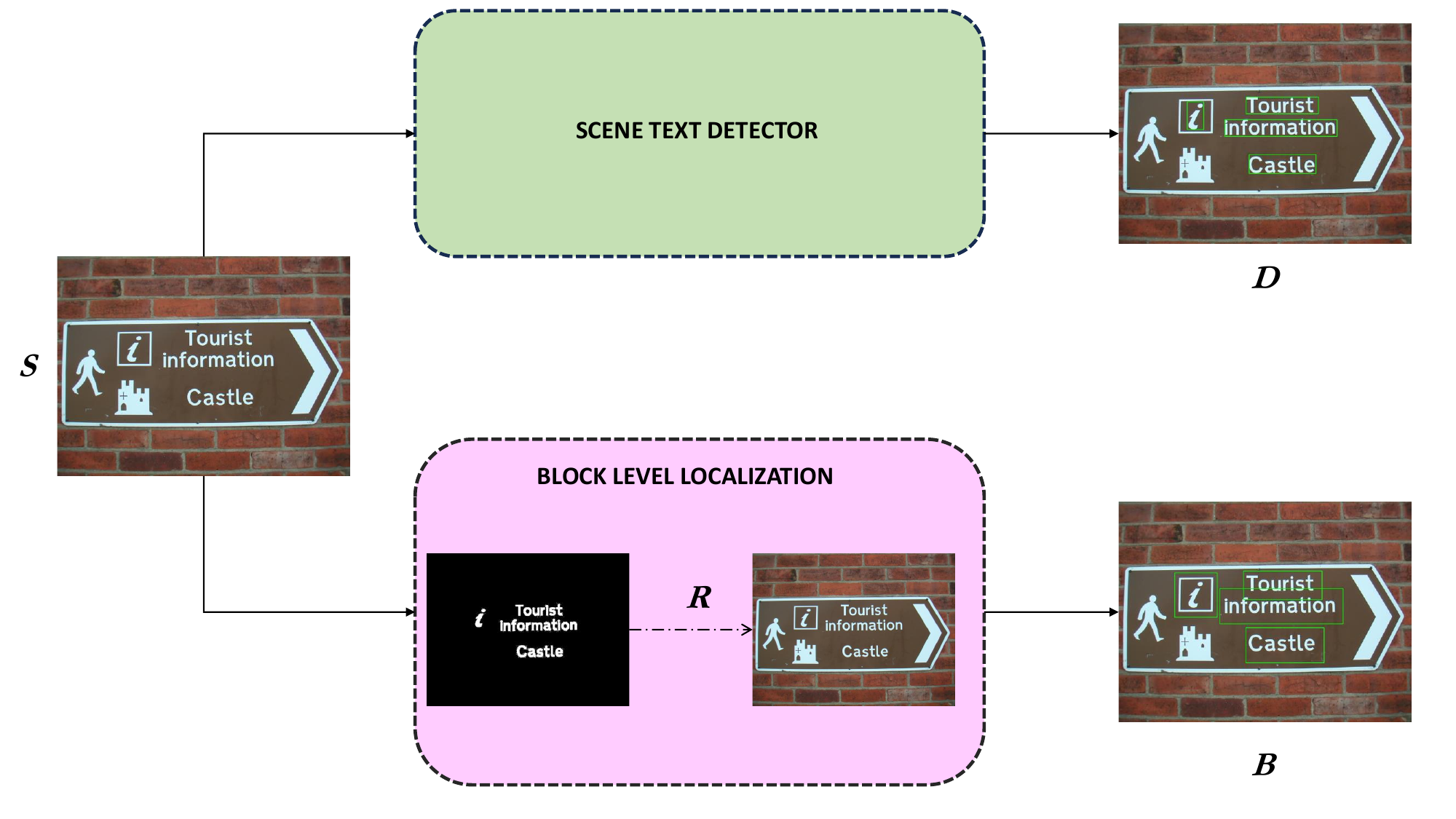}
    \caption{\textbf{A qualitative demonstration of traditional text detection against our proposed idea. }"S" refers to Source image, "D" refers to detector output, "B" refers to our block-based output and "R" refers to remapping to source image. The difference in localization in outputs is best viewed in zoom.}
    \label{fig:text_detection_impact}
    \vspace{-3pt}
\end{figure}
\textbf{Sensitivity to Scene Context.}  
To gauge caption quality, we re-ran ICDAR15 tests with: (i) descriptions removed and (ii) random, irrelevant descriptions. Table~\ref{tab:context_impact} shows that both settings increase DeepSolo fallbacks and false “confident” recognitions, with random captions harming performance most. Accurate, coherent scene text therefore remains critical.

\noindent\textbf{Skipping Detection.}  
Figure~\ref{fig:text_detection_impact} contrasts our relaxed, segmentation-based crops with detector-assisted boxes. Despite coarser localisation, recognition accuracy in Table \ref{tab:merged-recog-results} changes marginally, confirming that explicit, high-precision detection is often unnecessary. Eliminating it cuts compute while preserving state-of-the-art results.

\section{Conclusion}\label{sec: conclusion}
\begin{figure}[h]
    \centering
    \includegraphics[width=1\linewidth]{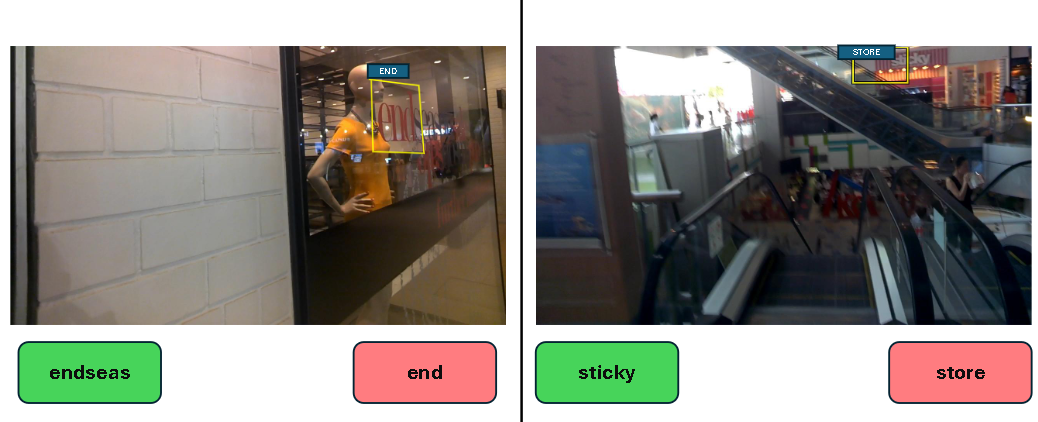}
    \caption{\textbf{Failure cases observed as false positives through our context-grounding metric. }(Green) denotes the target text, (Red) denotes the erroneous recognition result obtained after context-based scoring. Results here are shown for incidental scene text cases in ICDAR15 dataset.}
    \label{fig:failure_cases}
\end{figure}
We present a lightweight, context‐driven segmentation and recognition pipeline that reuses pretrained recognizers and captioners yet still encounters few key issues (failure examples in Fig.~\ref{fig:failure_cases}).  
(i) A single recognizer cannot read multiple scripts, so multilingual deployment needs separate models or explicit multilingual finetuning.  
(ii) Block-level localisation may fuse tightly packed words, passing segmentation errors to recognition.  
(iii) Inaccurate captions can bias the context scorer, boosting fallbacks to the heavier DeepSolo branch and eroding efficiency. Future work will (i) release multilingual recognizers, (ii) refine segmentation for dense layouts, and (iii) harden the pipeline against incidental scene text (orientation, lighting, occlusion) while retaining its speed advantage over end-to-end detectors.

\bibliographystyle{splncs04}
\bibliography{refs}

@misc{resnet50,
      title={Deep Residual Learning for Image Recognition}, 
      author={Kaiming He and Xiangyu Zhang and Shaoqing Ren and Jian Sun},
      year={2015},
      eprint={1512.03385},
      archivePrefix={arXiv},
      primaryClass={cs.CV},
      url={https://arxiv.org/abs/1512.03385}, 
}

@article{banerjee2024tts,
  title={TTS: Hilbert Transform-based Generative Adversarial Network for Tattoo and Scene Text Spotting},
  author={Banerjee, Ayan and Palaiahnakote, Shivakumara and Pal, Umapada and Antonacopoulos, Apostolos and Lu, Tong and Canet, Josep Llados},
  journal={IEEE Transactions on Multimedia},
  year={2024},
  publisher={IEEE}
}

@article{NDorder,
title = {NDOrder: Exploring a novel decoding order for scene text recognition},
journal = {Expert Systems with Applications},
volume = {249},
pages = {123771},
year = {2024},
issn = {0957-4174},
doi = {https://doi.org/10.1016/j.eswa.2024.123771},
url = {https://www.sciencedirect.com/science/article/pii/S0957417424006377},
author = {Dajian Zhong and Hongjian Zhan and Shujing Lyu and Cong Liu and Bing Yin and Palaiahnakote Shivakumara and Umapada Pal and Yue Lu}
}

@article{serialized,
author = {Ye, Qixiang and Doermann, David},
year = {2015},
month = {06},
pages = {},
title = {Text Detection and Recognition in Imagery: A Survey},
volume = {37},
journal = {IEEE Transactions on Pattern Analysis and Machine Intelligence},
doi = {10.1109/TPAMI.2014.2366765}
}

@inproceedings{batchnorm,
author = {Ioffe, Sergey and Szegedy, Christian},
title = {Batch normalization: accelerating deep network training by reducing internal covariate shift},
year = {2015},
publisher = {JMLR.org},
booktitle = {Proceedings of the 32nd International Conference on International Conference on Machine Learning - Volume 37},
pages = {448–456},
numpages = {9},
location = {Lille, France},
series = {ICML'15}
}

@article{ASTER,
author = {Shi, Baoguang and Yang, Mingkun and Wang, Xinggang and Lyu, Pengyuan and Yao, Cong and Bai, Xiang},
year = {2018},
month = {06},
pages = {1-1},
title = {ASTER: An Attentional Scene Text Recognizer with Flexible Rectification},
volume = {PP},
journal = {IEEE Transactions on Pattern Analysis and Machine Intelligence},
doi = {10.1109/TPAMI.2018.2848939}
}

@misc{xing2019convolutionalcharacternetworks,
      title={Convolutional Character Networks}, 
      author={Linjie Xing and Zhi Tian and Weilin Huang and Matthew R. Scott},
      year={2019},
      eprint={1910.07954},
      archivePrefix={arXiv},
      primaryClass={cs.CV},
      url={https://arxiv.org/abs/1910.07954}, 
}

@article{TextDragonAE,
  title={TextDragon: An End-to-End Framework for Arbitrary Shaped Text Spotting},
  author={Wei Feng and Wenhao He and Fei Yin and Xu-Yao Zhang and Cheng-Lin Liu},
  journal={2019 IEEE/CVF International Conference on Computer Vision (ICCV)},
  year={2019},
  pages={9075-9084},
  url={https://api.semanticscholar.org/CorpusID:207978383}
}

@InProceedings{synth800k,
  author       = "Ankush Gupta and Andrea Vedaldi and Andrew Zisserman",
  title        = "Synthetic Data for Text Localisation in Natural Images",
  booktitle    = "IEEE Conference on Computer Vision and Pattern Recognition",
  year         = "2016",
}

@inproceedings{SDTL,
  title={Synthetic data for text localisation in natural images},
  author={Gupta, Ankush and Vedaldi, Andrea and Zisserman, Andrew},
  booktitle={Proceedings of the IEEE conference on computer vision and pattern recognition},
  pages={2315--2324},
  year={2016}
}

@misc{vitaev2,
author = {Zhang, Qiming and Xu, Yufei and Zhang, Jing and Tao, Dacheng},
year = {2022},
month = {02},
pages = {},
title = {ViTAEv2: Vision Transformer Advanced by Exploring Inductive Bias for Image Recognition and Beyond},
doi = {10.48550/arXiv.2202.10108}
}

@inproceedings{deeptextspotter,
  title={Deep textspotter: An end-to-end trainable scene text localization and recognition framework},
  author={Busta, Michal and Neumann, Lukas and Matas, Jiri},
  booktitle={Proceedings of the IEEE international conference on computer vision},
  pages={2204--2212},
  year={2017}
}

@article{scut-ctw,
  title={Curved scene text detection via transverse and longitudinal sequence connection},
  author={Liu, Yuliang and Jin, Lianwen and Zhang, Shuaitao and Luo, Canjie and Zhang, Sheng},
  journal={Pattern Recognition},
  volume={90},
  pages={337--345},
  year={2019},
  publisher={Elsevier}
}

@misc{RARE,
      title={Robust Scene Text Recognition with Automatic Rectification}, 
      author={Baoguang Shi and Xinggang Wang and Pengyuan Lyu and Cong Yao and Xiang Bai},
      year={2016},
      eprint={1603.03915},
      archivePrefix={arXiv},
      primaryClass={cs.CV},
      url={https://arxiv.org/abs/1603.03915}, 
}

@INPROCEEDINGS{textocr,
  author={Singh, Amanpreet and Pang, Guan and Toh, Mandy and Huang, Jing and Galuba, Wojciech and Hassner, Tal},
  booktitle={2021 IEEE/CVF Conference on Computer Vision and Pattern Recognition (CVPR)}, 
  title={TextOCR: Towards large-scale end-to-end reasoning for arbitrary-shaped scene text}, 
  year={2021},
  volume={},
  number={},
  pages={8798-8808},
  doi={10.1109/CVPR46437.2021.00869}}

@INPROCEEDINGS{MLT-17,
  author={Nayef, Nibal and Yin, Fei and Bizid, Imen and Choi, Hyunsoo and Feng, Yuan and Karatzas, Dimosthenis and Luo, Zhenbo and Pal, Umapada and Rigaud, Christophe and Chazalon, Joseph and Khlif, Wafa and Luqman, Muhammad Muzzamil and Burie, Jean-Christophe and Liu, Cheng-lin and Ogier, Jean-Marc},
  booktitle={2017 14th IAPR International Conference on Document Analysis and Recognition (ICDAR)}, 
  title={ICDAR2017 Robust Reading Challenge on Multi-Lingual Scene Text Detection and Script Identification - RRC-MLT}, 
  year={2017},
  volume={01},
  number={},
  pages={1454-1459},
  doi={10.1109/ICDAR.2017.237}}

@inproceedings{MLT-19,
  title={ICDAR2019 robust reading challenge on multi-lingual scene text detection and recognition—RRC-MLT-2019},
  author={Nayef, Nibal and Patel, Yash and Busta, Michal and Chowdhury, Pinaki Nath and Karatzas, Dimosthenis and Khlif, Wafa and Matas, Jiri and Pal, Umapada and Burie, Jean-Christophe and Liu, Cheng-lin and others},
  booktitle={2019 International conference on document analysis and recognition (ICDAR)},
  pages={1582--1587},
  year={2019},
  organization={IEEE}
}

@misc{ctpn,
      title={Detecting Text in Natural Image with Connectionist Text Proposal Network}, 
      author={Zhi Tian and Weilin Huang and Tong He and Pan He and Yu Qiao},
      year={2016},
      eprint={1609.03605},
      archivePrefix={arXiv},
      primaryClass={cs.CV},
      url={https://arxiv.org/abs/1609.03605}, 
}

@ARTICLE{crnn,
  author={Shi, Baoguang and Bai, Xiang and Yao, Cong},
  journal={IEEE Transactions on Pattern Analysis and Machine Intelligence}, 
  title={An End-to-End Trainable Neural Network for Image-Based Sequence Recognition and Its Application to Scene Text Recognition}, 
  year={2017},
  volume={39},
  number={11},
  pages={2298-2304},
  doi={10.1109/TPAMI.2016.2646371}}

@article{kscomo,
  title={The Keywords Spotting with Context for Multi -Oriented Chinese Scene Text},
  author={Dao Wu and Rui Wang and Xiao‐Bo Tian and Dong Liang and Xiaochun Cao},
  journal={2018 IEEE Fourth International Conference on Multimedia Big Data (BigMM)},
  year={2018},
  pages={1-5},
  url={https://api.semanticscholar.org/CorpusID:53043043}
}

@inproceedings{textspotting-cnn2017,
  title={Towards end-to-end text spotting with convolutional recurrent neural networks},
  author={Li, Hui and Wang, Peng and Shen, Chunhua},
  booktitle={Proceedings of the IEEE international conference on computer vision},
  pages={5238--5246},
  year={2017}
}

@article{deep-learning,
author = {LeCun, Yann and Bengio, Y. and Hinton, Geoffrey},
year = {2015},
month = {05},
pages = {436-44},
title = {Deep Learning},
volume = {521},
journal = {Nature},
doi = {10.1038/nature14539}
}

@inproceedings{attention,
 author = {Vaswani, Ashish and Shazeer, Noam and Parmar, Niki and Uszkoreit, Jakob and Jones, Llion and Gomez, Aidan N and Kaiser, \L ukasz and Polosukhin, Illia},
 booktitle = {Advances in Neural Information Processing Systems},
 editor = {I. Guyon and U. Von Luxburg and S. Bengio and H. Wallach and R. Fergus and S. Vishwanathan and R. Garnett},
 pages = {},
 publisher = {Curran Associates, Inc.},
 title = {Attention is All you Need},
 url = {https://proceedings.neurips.cc/paper_files/paper/2017/file/3f5ee243547dee91fbd053c1c4a845aa-Paper.pdf},
 volume = {30},
 year = {2017}
}

@inproceedings{textfusenet,
  title     = {TextFuseNet: Scene Text Detection with Richer Fused Features},
  author    = {Ye, Jian and Chen, Zhe and Liu, Juhua and Du, Bo},
  booktitle = {Proceedings of the Twenty-Ninth International Joint Conference on
               Artificial Intelligence, {IJCAI-20}},
  publisher = {International Joint Conferences on Artificial Intelligence Organization},
  editor    = {Christian Bessiere},
  pages     = {516--522},
  year      = {2020},
  month     = {7},
  note      = {Main track},
  doi       = {10.24963/ijcai.2020/72},
  url       = {https://doi.org/10.24963/ijcai.2020/72},
}

@inproceedings{mpnet,
author = {Song, Kaitao and Tan, Xu and Qin, Tao and Lu, Jianfeng and Liu, Tie-Yan},
title = {MPNet: masked and permuted pre-training for language understanding},
year = {2020},
isbn = {9781713829546},
publisher = {Curran Associates Inc.},
address = {Red Hook, NY, USA},
booktitle = {Proceedings of the 34th International Conference on Neural Information Processing Systems},
articleno = {1414},
numpages = {11},
location = {Vancouver, BC, Canada},
series = {NIPS '20}
}

@inproceedings{leventshtein_ratio,
    title = "{JUNITMZ} at {S}em{E}val-2016 Task 1: Identifying Semantic Similarity Using {L}evenshtein Ratio",
    author = "Sarkar, Sandip  and
      Das, Dipankar  and
      Pakray, Partha  and
      Gelbukh, Alexander",
    editor = "Bethard, Steven  and
      Carpuat, Marine  and
      Cer, Daniel  and
      Jurgens, David  and
      Nakov, Preslav  and
      Zesch, Torsten",
    booktitle = "Proceedings of the 10th International Workshop on Semantic Evaluation ({S}em{E}val-2016)",
    month = jun,
    year = "2016",
    address = "San Diego, California",
    publisher = "Association for Computational Linguistics",
    url = "https://aclanthology.org/S16-1108/",
    doi = "10.18653/v1/S16-1108",
    pages = "702--705"
}

@inproceedings{icdar13-fst,
author = {Karatzas, Dimosthenis and Shafait, Faisal and Uchida, Seiichi and Iwamura, Masakazu and Bigorda, Lluis Gomez i. and Mestre, Sergi Robles and Mas, Joan and Mota, David Fernandez and Almaz\`{a}n, Jon Almaz\`{a}n and de las Heras, Llu\'{\i}s Pere},
title = {ICDAR 2013 Robust Reading Competition},
year = {2013},
isbn = {9780769549996},
publisher = {IEEE Computer Society},
address = {USA},
url = {https://doi.org/10.1109/ICDAR.2013.221},
doi = {10.1109/ICDAR.2013.221},
booktitle = {Proceedings of the 2013 12th International Conference on Document Analysis and Recognition},
pages = {1484–1493},
numpages = {10},
series = {ICDAR '13}
}

@inproceedings{Character_Region_Attention,
author = {Baek, Youngmin and Shin, Seung and Baek, Jeonghun and Park, Sungrae and Lee, Junyeop and Nam, Daehyun and Lee, Hwalsuk},
title = {Character Region Attention for Text Spotting},
year = {2020},
isbn = {978-3-030-58525-9},
publisher = {Springer-Verlag},
address = {Berlin, Heidelberg},
url = {https://doi.org/10.1007/978-3-030-58526-6_30},
doi = {10.1007/978-3-030-58526-6_30},
booktitle = {Computer Vision – ECCV 2020: 16th European Conference, Glasgow, UK, August 23–28, 2020, Proceedings, Part XXIX},
pages = {504–521},
numpages = {18},
location = {Glasgow, United Kingdom}
}

@misc{liu2018fotsfastorientedtext,
      title={FOTS: Fast Oriented Text Spotting with a Unified Network}, 
      author={Xuebo Liu and Ding Liang and Shi Yan and Dagui Chen and Yu Qiao and Junjie Yan},
      year={2018},
      eprint={1801.01671},
      archivePrefix={arXiv},
      primaryClass={cs.CV},
      url={https://arxiv.org/abs/1801.01671}, 
}

@INPROCEEDINGS{icdar15,
  author={Karatzas, Dimosthenis and Gomez-Bigorda, Lluis and Nicolaou, Anguelos and Ghosh, Suman and Bagdanov, Andrew and Iwamura, Masakazu and Matas, Jiri and Neumann, Lukas and Chandrasekhar, Vijay Ramaseshan and Lu, Shijian and Shafait, Faisal and Uchida, Seiichi and Valveny, Ernest},
  booktitle={2015 13th International Conference on Document Analysis and Recognition (ICDAR)}, 
  title={ICDAR 2015 competition on Robust Reading}, 
  year={2015},
  volume={},
  number={},
  pages={1156-1160},
  doi={10.1109/ICDAR.2015.7333942}}

@INPROCEEDINGS{totaltext,
  author={Ch'ng, Chee Kheng and Chan, Chee Seng},
  booktitle={2017 14th IAPR International Conference on Document Analysis and Recognition (ICDAR)}, 
  title={Total-Text: A Comprehensive Dataset for Scene Text Detection and Recognition}, 
  year={2017},
  volume={01},
  number={},
  pages={935-942},
  doi={10.1109/ICDAR.2017.157}}

@article{coco-ts,
  author       = {Simone Bonechi and
                  Paolo Andreini and
                  Monica Bianchini and
                  Franco Scarselli},
  title        = {COCO{\_}TS Dataset: Pixel-level Annotations Based on Weak Supervision
                  for Scene Text Segmentation},
  journal      = {CoRR},
  volume       = {abs/1904.00818},
  year         = {2019},
  url          = {http://arxiv.org/abs/1904.00818},
  eprinttype    = {arXiv},
  eprint       = {1904.00818},
  bibsource    = {dblp computer science bibliography, https://dblp.org}
}

@misc{masktextspotterv1,
      title={Mask TextSpotter: An End-to-End Trainable Neural Network for Spotting Text with Arbitrary Shapes}, 
      author={Pengyuan Lyu and Minghui Liao and Cong Yao and Wenhao Wu and Xiang Bai},
      year={2018},
      eprint={1807.02242},
      archivePrefix={arXiv},
      primaryClass={cs.CV},
      url={https://arxiv.org/abs/1807.02242}, 
}

@article{masktextspotterv2,
  author    = {Minghui Liao and Pengyuan Lyu and Minghang He and Cong Yao and Wenhao Wu and Xiang Bai},
  title     = {Mask TextSpotter: An end-to-end trainable neural network for spotting text with arbitrary shapes},
  journal   = {IEEE Transactions on Pattern Analysis and Machine Intelligence},
  year      = {2019},
  pages     = {11},
  doi       = {10.1109/TPAMI.2019.2937086}
}

@article{ADAM,
  title={Adam: A Method for Stochastic Optimization},
  author={Diederik P. Kingma and Jimmy Ba},
  journal={CoRR},
  year={2014},
  volume={abs/1412.6980},
  url={https://api.semanticscholar.org/CorpusID:6628106}
}

@INPROCEEDINGS{BCELOSS,
  author={Jadon, Shruti},
  booktitle={2020 IEEE Conference on Computational Intelligence in Bioinformatics and Computational Biology (CIBCB)}, 
  title={A survey of loss functions for semantic segmentation}, 
  year={2020},
  volume={},
  number={},
  pages={1-7},
  doi={10.1109/CIBCB48159.2020.9277638}}

@InProceedings{masktextspotterv3,
author={Liao, Minghui and Pang, Guan and Huang, Jing and Hassner, Tal and Bai, Xiang"},
title="Mask TextSpotter v3: Segmentation Proposal Network for Robust Scene Text Spotting",
booktitle="Computer Vision -- ECCV 2020",
year="2020",
publisher="Springer International Publishing",
address="Cham",
pages="706--722",
isbn="978-3-030-58621-8"
}

@inproceedings{craft,
  title={Character Region Awareness for Text Detection},
  author={Baek, Youngmin and Lee, Bado and Han, Dongyoon and Yun, Sangdoo and Lee, Hwalsuk},
  booktitle={Proceedings of the IEEE Conference on Computer Vision and Pattern Recognition},
  pages={9365--9374},
  year={2019}
}

@article{vitb,
  title={An Image is Worth 16x16 Words: Transformers for Image Recognition at Scale},
  author={Alexey Dosovitskiy and Lucas Beyer and Alexander Kolesnikov and Dirk Weissenborn and Xiaohua Zhai and Thomas Unterthiner and Mostafa Dehghani and Matthias Minderer and Georg Heigold and Sylvain Gelly and Jakob Uszkoreit and Neil Houlsby},
  journal={ArXiv},
  year={2020},
  volume={abs/2010.11929},
  url={https://api.semanticscholar.org/CorpusID:225039882}
}

@misc{he2018maskrcnn,
      title={Mask R-CNN}, 
      author={Kaiming He and Georgia Gkioxari and Piotr Dollár and Ross Girshick},
      year={2018},
      eprint={1703.06870},
      archivePrefix={arXiv},
      primaryClass={cs.CV},
      url={https://arxiv.org/abs/1703.06870}, 
}

@article{boundary,
author = {Hao, Wang and Lu, Pu and Zhang, Hui and Yang, Mingkun and Bai, Xiang and Xu, Yongchao and He, Mengchao and Wang, Yongpan and Liu, Wenyu},
year = {2020},
month = {04},
pages = {12160-12167},
title = {All You Need Is Boundary: Toward Arbitrary-Shaped Text Spotting},
volume = {34},
journal = {Proceedings of the AAAI Conference on Artificial Intelligence},
doi = {10.1609/aaai.v34i07.6896}
}

@article{textperceptron,
author = {Qiao, Liang and Tang, Sanli and Cheng, Zhanzhan and Xu, Yunlu and Niu, Yi and Pu, Shiliang and Wu, Fei},
year = {2020},
month = {04},
pages = {11899-11907},
title = {Text Perceptron: Towards End-to-End Arbitrary-Shaped Text Spotting},
volume = {34},
journal = {Proceedings of the AAAI Conference on Artificial Intelligence},
doi = {10.1609/aaai.v34i07.6864}
}

@article{pgnet,
author = {Wang, Pengfei and Zhang, Chengquan and Qi, Fei and Liu, Shanshan and Zhang, Xiaoqiang and Lyu, Pengyuan and Han, Junyu and Liu, Jingtuo and Ding, Errui and Shi, Guangming},
year = {2021},
month = {05},
pages = {2782-2790},
title = {PGNet: Real-time Arbitrarily-Shaped Text Spotting with Point Gathering Network},
volume = {35},
journal = {Proceedings of the AAAI Conference on Artificial Intelligence},
doi = {10.1609/aaai.v35i4.16383}
}

@article{mango,
author = {Qiao, Liang and Chen, Ying and Cheng, Zhanzhan and Xu, Yunlu and Niu, Yi and Pu, Shiliang and Wu, Fei},
year = {2021},
month = {05},
pages = {2467-2476},
title = {MANGO: A Mask Attention Guided One-Stage Scene Text Spotter},
volume = {35},
journal = {Proceedings of the AAAI Conference on Artificial Intelligence},
doi = {10.1609/aaai.v35i3.16348}
}

@ARTICLE{pan++,
  author={Wang, Wenhai and Xie, Enze and Li, Xiang and Liu, Xuebo and Liang, Ding and Yang, Zhibo and Lu, Tong and Shen, Chunhua},
  journal={IEEE Transactions on Pattern Analysis and Machine Intelligence}, 
  title={PAN++: Towards Efficient and Accurate End-to-End Spotting of Arbitrarily-Shaped Text}, 
  year={2022},
  volume={44},
  number={9},
  pages={5349-5367},
  doi={10.1109/TPAMI.2021.3077555}}

@InProceedings{Prasad_2018_ECCV,
author = {Prasad, Shitala and Kong, Adams Wai Kin},
title = {Using Object Information for Spotting Text},
booktitle = {Proceedings of the European Conference on Computer Vision (ECCV)},
month = {September},
year = {2018}
}

@inproceedings{background_cues_in_video,
author = {Wang, Lan and Wang, Yang and Shan, Susu and Su, Feng},
title = {Scene Text Detection and Tracking in Video with Background Cues},
year = {2018},
isbn = {9781450350464},
publisher = {Association for Computing Machinery},
address = {New York, NY, USA},
url = {https://doi.org/10.1145/3206025.3206051},
doi = {10.1145/3206025.3206051},
booktitle = {Proceedings of the 2018 ACM on International Conference on Multimedia Retrieval},
pages = {160–168},
numpages = {9},
location = {Yokohama, Japan},
series = {ICMR '18}
}

@manual{GoogleCloudVisionAPI,
  title        = {Cloud Vision API Documentation},
  author       = {{Google Cloud}},
  year         = 2025,
  url          = {https://cloud.google.com/vision/docs},
  note         = {Accessed: 2025-03-12}
}

@article{survey,
author = {Blanco-Medina, Pablo and Fidalgo, Eduardo and Alegre, Enrique and González-Castro, Víctor},
title = {A survey on methods, datasets and implementations for scene text spotting},
journal = {IET Image Processing},
volume = {16},
number = {13},
pages = {3426-3445},
doi = {https://doi.org/10.1049/ipr2.12574},
url = {https://ietresearch.onlinelibrary.wiley.com/doi/abs/10.1049/ipr2.12574},
eprint = {https://ietresearch.onlinelibrary.wiley.com/doi/pdf/10.1049/ipr2.12574},
year = {2022}
}

@INPROCEEDINGS{abcnet,
  author={Liu, Yuliang and Chen, Hao and Shen, Chunhua and He, Tong and Jin, Lianwen and Wang, Liangwei},
  booktitle={2020 IEEE/CVF Conference on Computer Vision and Pattern Recognition (CVPR)}, 
  title={ABCNet: Real-Time Scene Text Spotting With Adaptive Bezier-Curve Network}, 
  year={2020},
  volume={},
  number={},
  pages={9806-9815},
  doi={10.1109/CVPR42600.2020.00983}}

@ARTICLE{abcnetv2,
  author={Liu, Yuliang and Shen, Chunhua and Jin, Lianwen and He, Tong and Chen, Peng and Liu, Chongyu and Chen, Hao},
  journal={IEEE Transactions on Pattern Analysis and Machine Intelligence}, 
  title={ABCNet v2: Adaptive Bezier-Curve Network for Real-Time End-to-End Text Spotting}, 
  year={2022},
  volume={44},
  number={11},
  pages={8048-8064},
  doi={10.1109/TPAMI.2021.3107437}}

@article{testr,
  title={Text Spotting Transformers},
  author={Xiang Zhang and Yongwen Su and Subarna Tripathi and Zhuowen Tu},
  journal={2022 IEEE/CVF Conference on Computer Vision and Pattern Recognition (CVPR)},
  year={2022},
  pages={9509-9518},
  url={https://api.semanticscholar.org/CorpusID:247957876}
}

@InProceedings{detr,
author="Carion, Nicolas
and Massa, Francisco
and Synnaeve, Gabriel
and Usunier, Nicolas
and Kirillov, Alexander
and Zagoruyko, Sergey",
editor="Vedaldi, Andrea
and Bischof, Horst
and Brox, Thomas
and Frahm, Jan-Michael",
title="End-to-End Object Detection with Transformers",
booktitle="Computer Vision -- ECCV 2020",
year="2020",
publisher="Springer International Publishing",
address="Cham",
pages="213--229",
isbn="978-3-030-58452-8"
}

@article{swintextspotterv1,
  title = {SwinTextSpotter: Scene Text Spotting via Better Synergy between Text Detection and Text Recognition},
  author = {Mingxin Huang and YuLiang liu and Zhenghao Peng and Chongyu Liu and Dahua Lin and Shenggao Zhu and Nicholas Yuan and Kai Ding and Lianwen Jin},
  journal={arXiv preprint arXiv:2203.10209},
  year = {2022}
}

@misc{swintextspotterv2,
      title={SwinTextSpotter v2: Towards Better Synergy for Scene Text Spotting}, 
      author={Mingxin Huang and Dezhi Peng and Hongliang Li and Zhenghao Peng and Chongyu Liu and Dahua Lin and Yuliang Liu and Xiang Bai and Lianwen Jin},
      year={2024},
      eprint={2401.07641},
      archivePrefix={arXiv},
      primaryClass={cs.CV},
      url={https://arxiv.org/abs/2401.07641}, 
}

@inproceedings{TTS,
  title={Towards weakly-supervised text spotting using a multi-task transformer},
  author={Kittenplon, Yair and Lavi, Inbal and Fogel, Sharon and Bar, Yarin and Manmatha, R and Perona, Pietro},
  booktitle={Proceedings of the IEEE/CVF Conference on Computer Vision and Pattern Recognition},
  pages={4604--4613},
  year={2022}
}

@inproceedings{SPTS,
author = {Peng, Dezhi and Wang, Xinyu and Liu, Yuliang and Zhang, Jiaxin and Huang, Mingxin and Lai, Songxuan and Li, Jing and Zhu, Shenggao and Lin, Dahua and Shen, Chunhua and Bai, Xiang and Jin, Lianwen},
title = {SPTS: Single-Point Text Spotting},
year = {2022},
isbn = {9781450392037},
publisher = {Association for Computing Machinery},
address = {New York, NY, USA},
url = {https://doi.org/10.1145/3503161.3547942},
doi = {10.1145/3503161.3547942},
booktitle = {Proceedings of the 30th ACM International Conference on Multimedia},
pages = {4272–4281},
numpages = {10},
location = {Lisboa, Portugal},
series = {MM '22}
}

@misc{deepsolo,
author = {Ye, Maoyuan and Zhang, Jing and Zhao, Shanshan and Liu, Juhua and Liu, Tongliang and Du, Bo and Tao, Dacheng},
year = {2022},
month = {11},
pages = {},
title = {DeepSolo: Let Transformer Decoder with Explicit Points Solo for Text Spotting},
doi = {10.48550/arXiv.2211.10772}
}

@misc{deepsolo++,
author = {Ye, Maoyuan and Zhang, Jing and Zhao, Shanshan and Liu, Juhua and Liu, Tongliang and Du, Bo and Tao, Dacheng},
year = {2023},
month = {05},
pages = {},
title = {DeepSolo++: Let Transformer Decoder with Explicit Points Solo for Text Spotting},
doi = {10.48550/arXiv.2305.19957}
}

@article{str_in_the_wild,
author = {Chen, Xiaoxue and Jin, Lianwen and Zhu, Yuanzhi and Luo, Canjie and Wang, Tianwei},
title = {Text Recognition in the Wild: A Survey},
year = {2021},
issue_date = {March 2022},
publisher = {Association for Computing Machinery},
address = {New York, NY, USA},
volume = {54},
number = {2},
issn = {0360-0300},
url = {https://doi.org/10.1145/3440756},
doi = {10.1145/3440756},
journal = {ACM Comput. Surv.},
month = mar,
articleno = {42},
numpages = {35}
}

@article{textblockv2,
  title={TextBlockV2: Towards Precise-Detection-Free Scene Text Spotting with Pre-trained Language Model},
  author={Jiahao Lyu and Jin Wei and Gangyan Zeng and Zeng Li and Enze Xie and Wei Wang and Yu ZHOU},
  journal={ArXiv},
  year={2024},
  volume={abs/2403.10047},
  url={https://api.semanticscholar.org/CorpusID:268510467}
}

@InProceedings{parseq,
author="Bautista, Darwin
and Atienza, Rowel",
editor="Avidan, Shai
and Brostow, Gabriel
and Ciss{\'e}, Moustapha
and Farinella, Giovanni Maria
and Hassner, Tal",
title="Scene Text Recognition with Permuted Autoregressive Sequence Models",
booktitle="Computer Vision -- ECCV 2022",
year="2022",
publisher="Springer Nature Switzerland",
address="Cham",
pages="178--196",
isbn="978-3-031-19815-1"
}

@article{abinet,
  title={Read Like Humans: Autonomous, Bidirectional and Iterative Language Modeling for Scene Text Recognition},
  author={Shancheng Fang and Hongtao Xie and Yuxin Wang and Zhendong Mao and Yongdong Zhang},
  journal={2021 IEEE/CVF Conference on Computer Vision and Pattern Recognition (CVPR)},
  year={2021},
  pages={7094-7103},
  url={https://api.semanticscholar.org/CorpusID:232185272}
}

@inproceedings{vitstr,
author = {Atienza, Rowel},
title = {Vision Transformer for Fast and Efficient Scene Text Recognition},
year = {2021},
isbn = {978-3-030-86548-1},
publisher = {Springer-Verlag},
address = {Berlin, Heidelberg},
url = {https://doi.org/10.1007/978-3-030-86549-8_21},
doi = {10.1007/978-3-030-86549-8_21},
booktitle = {Document Analysis and Recognition – ICDAR 2021: 16th International Conference, Lausanne, Switzerland, September 5–10, 2021, Proceedings, Part I},
pages = {319–334},
numpages = {16},
location = {Lausanne, Switzerland}
}

@article{str_driving,
author = {Zhang, Chongsheng and Yuefeng, Tao and Du, Kai and Ding, Weiping and Wang, Bin and Liu, Ji and Wang, Wei},
year = {2021},
month = {09},
pages = {1-1},
title = {Character-level Street View Text Spotting Based on Deep Multi-Segmentation Network for Smarter Autonomous Driving},
volume = {PP},
journal = {IEEE Transactions on Artificial Intelligence},
doi = {10.1109/TAI.2021.3116216}
}

@ARTICLE{str_robot_navigation,
  author={Desouza, G.N. and Kak, A.C.},
  journal={IEEE Transactions on Pattern Analysis and Machine Intelligence}, 
  title={Vision for mobile robot navigation: a survey}, 
  year={2002},
  volume={24},
  number={2},
  pages={237-267},
  doi={10.1109/34.982903}}

@misc{str_the_deep_learning_era,
author = {Long, Shangbang and He, Xin and Ya, Cong},
year = {2018},
month = {11},
pages = {},
title = {Scene Text Detection and Recognition: The Deep Learning Era},
doi = {10.48550/arXiv.1811.04256}
}

@inproceedings{BLIP2,
author = {Li, Junnan and Li, Dongxu and Savarese, Silvio and Hoi, Steven},
title = {BLIP-2: bootstrapping language-image pre-training with frozen image encoders and large language models},
year = {2023},
publisher = {JMLR.org},
booktitle = {Proceedings of the 40th International Conference on Machine Learning},
articleno = {814},
numpages = {13},
location = {Honolulu, Hawaii, USA},
series = {ICML'23}
}

@InProceedings{unet,
author="Ronneberger, Olaf
and Fischer, Philipp
and Brox, Thomas",
editor="Navab, Nassir
and Hornegger, Joachim
and Wells, William M.
and Frangi, Alejandro F.",
title="U-Net: Convolutional Networks for Biomedical Image Segmentation",
booktitle="Medical Image Computing and Computer-Assisted Intervention -- MICCAI 2015",
year="2015",
publisher="Springer International Publishing",
address="Cham",
pages="234--241",
isbn="978-3-319-24574-4"
}

@article{text_segmentation_survey,
  title={Text segmentation techniques: a critical review},
  author={Pak, Irina and Teh, Phoey Lee},
  journal={Innovative Computing, Optimization and Its Applications: Modelling and Simulations},
  pages={167--181},
  year={2018},
  publisher={Springer}
}

@InProceedings{deeplabv3+,
author="Chen, Liang-Chieh
and Zhu, Yukun
and Papandreou, George
and Schroff, Florian
and Adam, Hartwig",
editor="Ferrari, Vittorio
and Hebert, Martial
and Sminchisescu, Cristian
and Weiss, Yair",
title="Encoder-Decoder with Atrous Separable Convolution for Semantic Image Segmentation",
booktitle="Computer Vision -- ECCV 2018",
year="2018",
publisher="Springer International Publishing",
address="Cham",
pages="833--851",
isbn="978-3-030-01234-2"
}

@INPROCEEDINGS{textformer,
  author={Wang, Xiaocong and Wu, Chaoyue and Yu, Haiyang and Li, Bin and Xue, Xiangyang},
  booktitle={2023 IEEE International Conference on Multimedia and Expo (ICME)}, 
  title={TextFormer: Component-aware Text Segmentation with Transformer}, 
  year={2023},
  volume={},
  number={},
  pages={1877-1882},
  doi={10.1109/ICME55011.2023.00322}}

\end{document}